\documentclass{article}

\usepackage{PRIMEarxiv}
\usepackage{tabularx}
\usepackage[utf8]{inputenc} % allow utf-8 input
\usepackage[T1]{fontenc}    % use 8-bit T1 fonts
\usepackage{hyperref}       % hyperlinks
\usepackage{url}            % simple URL typesetting
\usepackage{booktabs}       % professional-quality tables
\usepackage{amsfonts}       % blackboard math symbols
\usepackage{nicefrac}       % compact symbols for 1/2, etc.
\usepackage{microtype}      % microtypography
\usepackage{lipsum}
\usepackage{fancyhdr}       % header
\usepackage{graphicx}       % graphics
\usepackage[labelfont=bf]{caption}
\usepackage{threeparttable}
\usepackage{subcaption}
\usepackage{enumitem}
\usepackage{adjustbox}
\graphicspath{{media/}}     % organize your images and other figures under media/ folder

\title{Synthetic Text Detection: Systemic Literature Review }
\author{
  Jesus Guerrero \\
  Texas A\&M University - San Antonio\\
  San Antonio\\
  \texttt{jguer017@jaguar.tamu.edu} \\
   \And
  Izzat Alsmadi \\
  Texas A\&M University - San Antonio\\
  San Antonio\\
  \texttt{Izzat.Alsmadi@tamusa.edu} \\
}

\begin{document}
\maketitle

\begin{abstract}
Within the text analysis and processing fields, generated text attacks have been made easier to create than ever before. To combat these attacks open sourcing models and datasets have become a major trend to create automated detection algorithms in defense of authenticity. For this purpose, synthetic text detection has become an increasingly viable topic of research. This review is written for the purpose of creating a snapshot of the state of current literature and easing the barrier to entry for future authors. Towards that goal, we identified few research trends and challenges in this field, 
\end{abstract}

% keywords can be removed
%\keywords{}

\section{Introduction}
 Studies regarding text generation before 2017 were generally scarce and far between. As the body of research grew for synthetic text generation, so did the research for detectors, though lagging behind. 

This paper discusses current trends of research and future viable research options with the goal of shortening the research process for Artificial Intelligence generated text detection. The main topic of discussion is the literature itself using the PRISMA methodology. Detection literature was reviewed systematically and put together in detail for novel research. 

\subsection{Related Surveys}
There are seven surveys near the topic of synthetic text detection. Seven \cite{Fatima2022f, Alsmadi2022, Jawahar2020e, Dong2021e, Li2021f, Iqbal2022, Celikyilmaz2020} are about reviewing the literature on the generation process whereby current techniques, domains, data sets and models are shown and reviewed as available. Those seven surveys are very useful for discovering where the body of research is today in regards to text generation. 

In the current time very few detection based reviews and surveys exist, with only \cite{Jawahar2020e} truly narrowing itself down to actual detection. This survey on binary classification is a valuable contribution to the state of detecting fake text. This article shows techniques and models commonly used in 2020 and since then there have been more relevant publications than ever before.  

\subsection{Main contributions}
At the time of the previous surveys/reviews the body of research was perhaps too small for a worthwhile review of primary sources. Since then many of the techniques, models and data sets have become more accessible and as a defense against attack, open-source. Now in the year 2022 we can update and add to these related surveys. For this literature review we have these main contributions: 
\begin{itemize}
\item A review of 50 related articles about synthetic text detection 
\item Shows recent innovations for detection. 
\item Shows gaps in current research for future work. 
\end{itemize}
This is perhaps one of the first literature reviews on the narrow topic of generated text detection. To the best of our knowledge there have been no systemic literature reviews on detecting synthetic text. This study focuses on exploring the current research literature, showing the current ecosystem behind synthetic detection and preparing for future research. 

\section{Research Design}
For our systematic literature review, we used current research tools to aid in following the systemic process, PRISMA. The research design here is to find and compile the most relevant body of research for distinguishing fake text and making inquiries. We setup 3 research questions, followed the review process and distilled research to include in the literature review. There were stages of collecting the articles involved, starting with collecting many articles by title then following an exclusion/inclusion process down to 50 papers. 

For the actual searching itself the main search engine used was Google Scholar as it is an aggregate of other databases and engines. The publisher, article type and year was recorded and notated in a 3rd party app, Mendeley. 1,211 related articles were chosen from their titles on Google Scholar using specified keywords, approved by a supervising author.  

The articles were scrapped from the Google Scholar website using Mendeley’s browser extension scrapper and were automatically added and kept in a database of Mendeley’s new reference manager to speed up the process. The collections feature of the application was used to separate the stages of the PRISMA methodology.  

\subsection{Research Questions}
With the unifying goal of preparing for future research, the inclusion/exclusion process included more and more study of the given texts. These research questions and research objectives were created to guide the process of choosing articles and scrutinizing the literature for that purpose: 

\textbf{{Table 1}}: Research Questions 
\begin{table}[h]
\begin{tabular}{|p{1.26cm}p{15.23cm}|}
\hline
\multicolumn{2}{|c|}{\textbf{Research Question}}                                                                     \\ \hline
\multicolumn{1}{|l|}{\textbf{RQ1}} & Which datasets are currently used in the literature to detect deep fake models? \\ \hline
\multicolumn{1}{|l|}{\textbf{RQ2}} & What accuracy evaluation methods are there for detection effectiveness?         \\ \hline
\multicolumn{1}{|l|}{\textbf{RQ3}} & What impact have recent innovations had on fake text detection?                 \\ \hline
\end{tabular}
\end{table}

\subsection{Research Objectives}
For this study there are 5 objectives: 
\begin{itemize}
\item To investigate the current existing techniques/approaches of detecting artificial text. 
\item To explore models and datasets created to detect artificial text. 
\item To explore accuracy evaluation of fake text detection. 
\item Show recent innovations since previous surveys. 
\item Show future work for further research.  
\end{itemize}

\subsection{Searching Strategy To Retrieve Studies}
Given most to all studies were queried on Google Scholar with keywords including text generation, detection and synthetic text, the engine includes searches into various other databases such as IEEE, ArXiv, Semantic Scholar, Springer, ACM journals, Elsevier and more, even including schools inside one search page. 
Keywords were used in regards to their category. Specifically in this SLR each set of keywords contributed a number of articles but some were more valuable than others. “Text generation detection” was perhaps the most fruitful, though, the body of research is quite small. A wide variety of query keywords had to be used to gather the largest possible pool of articles regarding text generation. For a few cases the title did not appear to be about synthetic detection, though upon further reading was in fact relevant to detection and vice versa. 

\newpage

\textbf{{Table 2}}: Keywords used by category
\begin{table}[h]
\begin{tabular}{|l|l|l|l|l|}
\hline
\multicolumn{1}{|c|}{Domain}  & Text generation method                & Sample size & Text generation innovations & Classifier     \\ \hline
\multicolumn{1}{|l|}{} &                       & Large       & Natural Language Processing &                     \\
Social Networks               & GANs & Small       & Text analysis               & Word embedding       \\
Fake news                     & Fake text                             & Sample      & Text classification         & CNN \\
Domain                        & Augmentation                          & Training    &                             & RNN     \\
Low resource                  &                                       & Models      &                             & Transformers                 \\
                              &                                       &             &                             & Detection                    \\
                              &                                       &             &                             & LSTM       \\
                              &                                       &             &                             & Ensamble                     \\ \hline
\end{tabular}
\end{table}

From these keywords they were assorted with AND, OR and quotation required clauses. Certain key words like “text generation” AND “language processing” were especially effective together at finding articles while “fake text” led to irrelevant topics. Though some keywords were only useful for finding niche articles. 

\subsection{Article Inclusion Exclusion Criteria}
A total of 1,211 articles were found using the above queries. With duplicates removed, the remaining 1,041 articles were sifted. Many articles containing relevant keywords and titles were not about text generation, Natural Language Processing (NLP) and detection.\\
Some articles were not machine centric and were about societal or human reading of generated text though they were about generated text detection. Others mentioned fake text as trolling, which is not the focus of this review. 

In the partial review many of these papers were excluded by abstract because they were not machine centric or were based on societal differences. Out of the 1,041 articles after removing duplicates, a partial review left 381 articles eligible for full review. The following criteria was used to include or exclude these papers from this point on: 

The following is the inclusion criteria: 
\begin{itemize}
\item The article must include machine generated text classification or be highly relevant 
\item The article can include other languages in its dataset
\item The article itself must be written in English
\item Surveys on text generation are allowed 
\end{itemize}

The following is the exclusion criteria: 
\begin{itemize}
\item The models used must be machine-centric, meaning they require machine learning to determine if a sample is generated text.
\end{itemize}

\section{Systematic Mapping Study Results}
Here we show the results of the systematic study, stages of the inclusion/exclusion process,  publisher names and dates of publication. This overviews the current status of synthetic text detection literature in late 2022 and records potential research gaps to be filled by future authors.

\newpage

\begin{figure}[h]
    \captionsetup{singlelinecheck = false, format= hang, justification=raggedright, labelsep=space}
    \centering
    \begin{measuredfigure}
        \caption{Inclusion exclusion process }
        \includegraphics[scale=0.75]{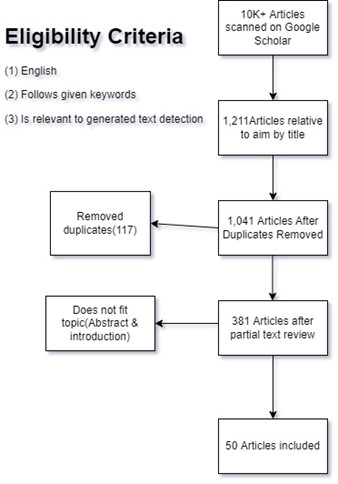}              
    \end{measuredfigure}
\end{figure}
 
\begin{figure}[h]
\centering
\begin{minipage}{.5\textwidth}
  \centering
  \includegraphics[width=.85\linewidth]{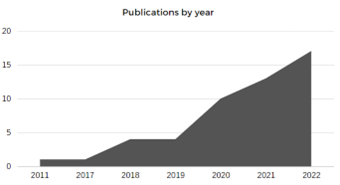}
  \captionof{figure}{Publications by year}
  \label{fig:test1}
\end{minipage}%
\begin{minipage}{.5\textwidth}
  \centering
  \includegraphics[width=.85\linewidth]{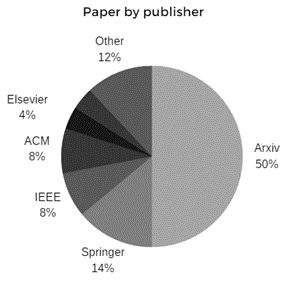}
  \captionof{figure}{Articles by publisher}
  \label{fig:test2}
\end{minipage}
\end{figure}

\begin{adjustbox}{margin=0 9 0 0}
\textbf{RQ1:} \textit{Which datasets/models are currently used in the literature to detect generated text?}
\end{adjustbox}\\
According to literature, the more training data for both human and machine generated text the better the outcome will become. Though there are many sources for both real and fake it is good to see the popular ones for specific domains. Below are datasets usable for training a model for actual detection and research in the time to come: 

\textit{Open-source Datasets}
\begin{itemize}[label=$\circ$]
\item Hugging face: \url{https://huggingface.co/datasets} \\
This website is the first place to look for datasets and models. Hugging face has a plethora to choose from across many regards of machine learning. Most to all of those below can be found on the platform.
\item GPT-3: \url{https://openai.com/api/} \\
This is a high quality source of generated text. There are several models to choose from for GPT-3 though the models are not free to use. 
\item GPT-2: \url{https://github.com/openai/gpt-2-output-dataset/ } \\
For a while this dataset was standard in its use for text generation. This dataset includes samples of both synthetic and real text. 
\item Grover: \url{https://github.com/rowanz/grover} \\
This is more of a collection of scripts for making a dataset oriented for news articles. The repo includes a detection model, text generator, accuracy evaluator and a web crawler for gathering authentic text source data.  
\item Authorship Attribution: \url{https://bit.ly/3DNlLxw} \\
This is a dataset for detecting specific popular text generators. The csv samples for these generators are placed in the above link. The focus is more based on news/political articles. 
\item TuringBench: \url{https://github.com/TuringBench/TuringBench } \\
The main website for TuringBench is more of a leaderboard about detecting which generator is being used. The dataset is given in a zip file and whoever gets the highest accuracy rating for detecting the generator wins. 
\item Academia papers: \url{https://github.com/vijini/GeneratedTextDetection/tree/main/Dataset } \\
A niche dataset for synthetic academia papers, though, it is small and is not condensed in one file. This would be good to expand upon as a separate research paper. 
\item TweepFake: \url{https://github.com/tizfa/tweepfake_deepfake_text_detection/} \\
A popular twitter dataset with human and machine tweets
\end{itemize}

\textit{Open source generative models }
\begin{itemize}[label=$\circ$]
\item Grover: \url{https://grover.allenai.org/} 
\item GPT-2: \url{https://github.com/openai/gpt-2} 
\item GPT-3 group(Paid): \url{https://openai.com/api/} 
\item Hugging face: \url{https://bit.ly/3LwGszE} \\
I mention this again, with over 5,000 models to choose from, you can see the limitations of text generation, GPT-2 being the most popular on the platform.
\item Web app w/ text generation (GPT-2), (Grover): \url{https://app.inferkit.com/demo} 
\end{itemize}

\begin{adjustbox}{margin=0 9 0 0}
\textit{Existing detective models}
\end{adjustbox}\\
In late 2022 detectors usually are rarer, have to be trained with a generated text dataset, and are not often pre-built. Though new models and prebuilt detectors are being created all the time and now popping up more rapidly. There are likely many on the HuggingFace platform for example. Some models also serve a dual usage, having both detector and generator.  

Below, an example of a pre-trained model, BERT, is good for general text classification modeling,  requiring further building to fully classify generated text as real or fake. GLTR is another example model, a human detection helper which improves human-centric detection. GLTR colors words which are most suspiciously generated, boosting accuracy quite a bit with minimal learning from the person. The rest below are pre-built and available for detection testing: 
\begin{itemize}[label=$\circ$]
\item BERT based modeling 
\item GLTR: \url{gltr.io }
\item Grover: \url{https://grover.allenai.org/}(2019) 
\item Open-AI GPT-2: \url{https://huggingface.co/openai-detector/} (2019) 
\item RoFT: \url{https://roft.io/} (gamification of human detection) 
\end{itemize}

\begin{adjustbox}{margin=0 9 0 0}
\textbf{RQ2:} \textit{What accuracy evaluation methods are there for detection effectiveness?}
\end{adjustbox}\\
According to \cite{Li2022} a good general rule of model evaluation is testing the mislabel or error rate. This can mean testing against a given test/validation set or testing against an outside dataset. Using a recorded error rate you can also distinguish how effective a detector is per generative model.  

The standard way most detectors are evaluated in the literature is to stay in the same domain and use similar datasets. The error rates are based on a narrow binary classification and due to this there are much better error rates. You can see this throughout the literature where there is one domain, like TweepFake\cite{FagniFabrizio;GambiniMargherita;MartellaAntonio;TesconiMaurizio2021e}, news oriented models \cite{Zellers2019}, language based models \cite{Chen2022f}, and other niche categories sticking to their own evaluated domains.  

To truly test a model against more real data here are two things done in the literature; adversarial testing \cite{Wolff2020} and sourcing different generators \cite{Li2022}.  For adversarial testing this means accounting for a post-processing phase of text generation whereby Greek and uncommon symbols are added along with misspellings and surely other techniques in adversarial fashion to cause the detector to mislabel the text as human written. Testing against this noise is a very good way to evaluate a real-world model. A typical solution for adversarial attack is with a preprocessing phase for the detector. 

For evaluating accuracy against different sources, a good idea from \cite{Li2022} is to record the error rates of different common and popular generator. Below is an example of how a detector can mislabel a text, 

\begin{figure}[h]
    \captionsetup{singlelinecheck = false, format= hang, justification=raggedright, labelsep=space}
    \centering
    \begin{measuredfigure}
        \caption{Example evaluation based on source, \cite{liartificial}}
        \includegraphics[scale=0.8]{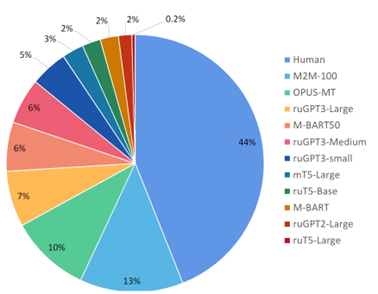}              
    \end{measuredfigure}
\end{figure}

The authors of the article found human text was most mislabeled as synthetic. This can be used to adjust a model and evaluate its accuracy and can further be divided by which sources the human written text originated. Though a major flaw later stated by the author is how common it is to fine tune a generator, making labelling all generators unfeasible. 

Lastly, how a model will be used should always be part of the evaluation. Low resource training may be better for something like TweepFake, captioning or microblog detection. Though general language detection would require more resources and vary in effectiveness based on the text in question plus how similar the training set had been. So the evaluation can be divided up in some cases or narrowed in others. This way the evaluation is accurate to the model’s purpose. 

\begin{adjustbox}{margin=0 9 0 0}
\textbf{RQ3:} \textit{What impact have recent innovations had on fake text detection?}
\end{adjustbox}\\
Across the literature the general setting of recent innovations is to combat fake text, detecting and removing its effect on public discourse. Paid models like GPT-3 are generally superior at creating advanced generated text for fooling human detection but is made relatively open source for research against itself and other high quality generators. 

There is a trend to open source datasets and models to protect the community from synthetic text by making it easier to create both generation and detection models. These motivations perhaps have had the greatest impact on future research. Many articles, as seen in Figure 2; publications by year, have been written more recently. Synthetic text will likely be more highly researched in the next few years. 

Particularly niche and low resource domains are being filled in with working models and novel solutions. Short and long form AI generated document detection models are now more numerous with their accuracy reported in their respective papers. These new models give us more options than the standard detector, opening up more difficult and niche types. 

\section{Identified research gaps}

Here we discuss the limitations of current research. The gaps are separated by 5 aspects of synthetic text detection gathered from our research questions. From these gaps future research can be studied. 

\begin{enumerate}
  \item \textbf{Limited overall research.} As of mid 2022 there is a scarcity of research papers regarding artificial text detection. The majority of time spent creating this SLR was in curating as many articles as possible and even with that time spent there were still a relatively few papers. 
  \item \textbf{Limited research on adversarial attacks.} Pre and post processing methodologies are missing for attacking detectors. Here is one example on adversarial attacking\cite{Wolff2020}. 
  \item \textbf{Limited evaluation methodologies for detectors.} Several papers existed for evaluation methods though nothing thorough. For RQ2 the information was pulled from small parts of a group of papers but there was not much research outside of that. 
  \item \textbf{Low resource detector optimization.} Low resource training also had limited research. TweepFake\cite{FagniFabrizio;GambiniMargherita;MartellaAntonio;TesconiMaurizio2021e} and fake academic paper detection\cite{Liyanage2022f} were perhaps the most optimization related articles. There is a gap here.
  \item \textbf{Research in other languages.} Most other languages have very limited research though some articles do exist\cite{Chen2022f}, \cite{skrylnikovartificial}, \cite{Shamardina2022}. Data sets exist for Chinese, Russian and other languages as well but very few synthetic language detectors outside English. 
\end{enumerate}

\section{Recommendations and future research directions}

There is plenty of room for research in AI text detection across different aspects. Given the limited overall research the field can be taken from many angles. This includes studies on increasing accuracy for specific domains such as news, blogs, social media outlets, books, academia. Most to all domains are open for detection modeling. 

Language specific research is a definite direction a person can take. Spanish, Hindu and many others do not have synthetic text detection research. Remaking previous research paper detectors in different languages is a good bet as well as dataset creation for future authors. There is little to no research on low resource generated text detection languages and domains as well. 

Better, more robust, evaluation methods for detectors is a potential topic. Tasks like generalized accuracy tests or post/pre processing methodologies are open game. In this vein of evaluation adversarial detector attacks are a great and open avenue for research. In \cite{Wolff2020} adding simple homoglyphs break most detectors and can easily be added to generation models. Misspelling also helps in adversarial text generation. Together the detector recall goes from 97\% down to 0.26\%, massively fooling detection 

In addition there are many niche domains where text generation and even some detectors exist but there is no adversarial research behind the domain. A paper like \cite{Le2020} takes an adversarial approach to generating synthetic comments to fool detectors. To this end only a handful of research was found. 

Another open but more difficult research niche is authorship attribution. Not just binary detection but multi-label detection for the most popular models. An example of this research is previously mentioned TuringBench, whereby an online leaderboard was created to detect which model the generator is sourced. 

\section{Conclusion}

Natural language processing is trending in good fashion with plenty of open source projects and ideas for novelty. Automated AI text generation has grown tremendously in the past four years and now there is a fresh need for detection. As we enter a phase of defending trolls, bots and generated commentary recent advancements have allowed for more of itself. In this paper, our focus is on synthetic text generation with possible research trends and challenges.  

%Bibliography
\nocite{*}
\bibliographystyle{plain}  
\bibliography{references}

\end{document}